%% file: main.tex
\documentclass[11pt]{article}
\pdfoutput=1

\usepackage{ACL2023}

\usepackage{times}
\usepackage{latexsym}
\usepackage{booktabs}
\usepackage{graphicx}
\usepackage{amsmath}
\usepackage{pgfplots}
\usepackage{algorithm} 
\usepackage{hyperref}
\usepackage{algpseudocode}
\usepackage{algorithmicx}

\usepackage[T1]{fontenc}

\usepackage[utf8]{inputenc}

\usepackage{microtype}

\usepackage{inconsolata}

\title{Adam-Smith at SemEval-2023 Task 4: Discovering Human Values in Arguments with Ensembles of Transformer-based Models}

\author{Daniel Schroter, Daryna Dementieva \and Georg Groh \\
  TU Munich, Department of Informatics, Germany \\
  \texttt{\texttt{daniel@schroter.biz}, \texttt{daryna.dementieva@tum.de}, \texttt{grohg@in.tum.de}}}

\begin{document}
\maketitle
\begin{abstract}
This paper presents the best-performing approach alias "Adam Smith" for the SemEval-2023 Task 4: "Identification of Human Values behind Arguments". The goal of the task was to create systems that automatically identify the values within textual arguments. We train transformer-based models until they reach their loss minimum or f1-score maximum. Ensembling the models by selecting one global decision threshold that maximizes the f1-score leads to the best-performing system in the competition. Ensembling based on stacking with logistic regressions shows the best performance on an additional dataset provided to evaluate the robustness ("Nahj al-Balagha"). Apart from outlining the submitted system, we demonstrate that the use of the large ensemble model is not necessary and that the system size can be significantly reduced. 
\end{abstract}

\section{Introduction}
"We should ban whaling, whaling is wiping out species for little in return." say some, "We shouldn't, it is part of a great number of cultures." say others. Both arguments support their claim, but why are some arguments more convincing to us than others? This might relate to the underlying values they address. Whereas the first argument appeals to the value of "universalism: nature", the second one addresses the value of "tradition". Whether an argument is in agreement or disagreement with our values influences its ability to persuade. The task organizers \cite{kiesel:2023} are the first who extend the field of argument mining by this "value" dimension. As part of the SemEval-2023 workshop, they organize the task of automatically detecting human values behind arguments. They decided to add two additional test datasets to evaluate the robustness of the developed systems \cite{mirzakhmedova:2023}.

Our system uses an ensemble of transformer-based models that are either trained until they reach their loss minimum or the f1-score maximum. To ensemble the models we average the individual predictions and calculate a decision threshold for the final system on a separate "Leave-out-Dataset". \textbf{This model achieves the best performance in the competition} ("Main" dataset). Each team was allowed to submit up to four systems. Ensembling the predictions by using stacking with logistic regressions leads to the system with the best performance on the additional "Nahj al-Balagha" dataset. In this paper, we describe the best-performing system on the "Main" dataset and briefly outline the ideas behind the other submitted systems. 

The system can be accessed through a web-based interface that is available online\footnote{\url{https://values.args.me/}}. Furthermore, a docker container, models, and code are open source and publicly accessible (Appendix \ref{app:experiments}).

\section{Background}\label{sec:background}

\citet{mirzakhmedova:2023} created a labeled dataset of 9324 arguments from 6 different sources. The arguments are structured as follows: 

\input{tables/example.tex}

 The arguments are in English and in total there are 20 different value categories to predict. Each argument is labeled with one or multiple values. Hence the task at hand can be characterized as a multi-labeling problem. The systems can be tested against two additional datasets to evaluate their robustness on unseen data from different domains. The "Nahj al-Balagha" dataset contains arguments from Islamic religious texts. The "New York Times" dataset contains arguments from texts about COVID-19. A detailed introduction to the datasets can be found in \citet{mirzakhmedova:2023}. The task organizers \cite{kiesel:2023} created two baseline models: 1-baseline and a BERT-based system. The system we propose builds upon the transformer architecture by \citet{vaswani2017attention} and in particular the BERT model \cite{Devlin:2018}. In fact, we use two improved versions of the original BERT model called RoBERTa \cite{liu2019roberta} and DeBERTa \cite{he2021deberta}. Further, we apply well-known techniques from the field of practical AI such as ensembling \cite{zhou2012ensemble}, cross-validation and early-stopping \cite{GoodBengCour16}. By presenting the best-performing system in the task and outperforming the baselines by a large margin, this paper delivers valuable insights into how values can be automatically detected within text.

\section{System Overview}
\begin{figure}[t]
    \centering
    \includegraphics[width=\linewidth]{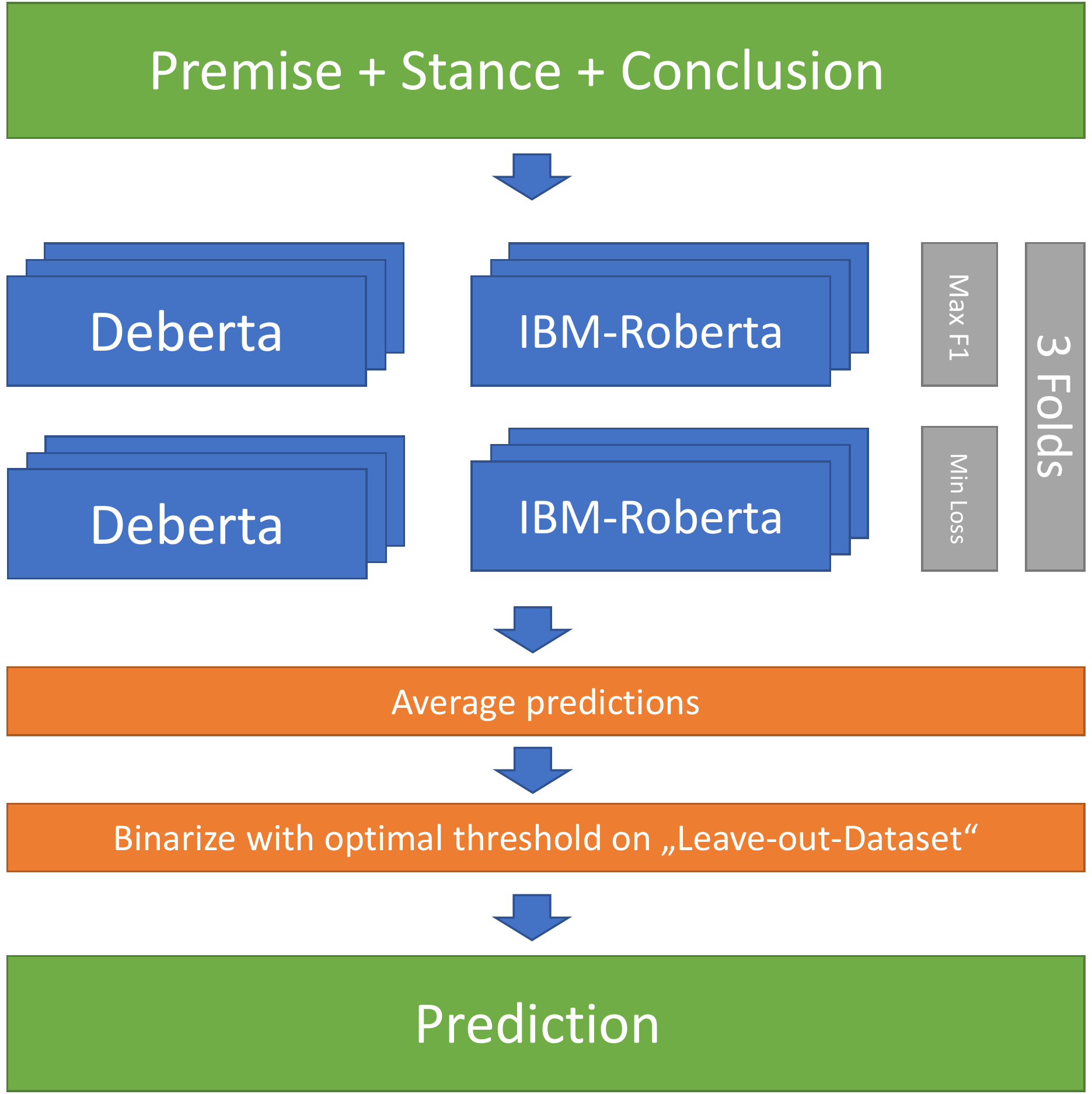}\hfill
    \caption{Inference System}
    \label{fig:architecture}
\end{figure}

This section is dedicated to precisely describing the best-performing submission as a baseline for future research and further development. The proposed system is an ensemble of 12 individual models. Figure \ref{fig:architecture} provides an overview of the inference pipeline of the final system. We briefly outline the process of making a prediction and subsequently describe each of the individual steps in detail. To make a prediction the following steps are performed: 

1. We take an input argument and concatenate the premise, stance, and conclusion.

2. The input is then fed into the neural networks. The output of each neural network is a vector containing 20 values with the "confidence" (values between 0 and 1) whether the sample has the corresponding label. The final system consists of 12 models, so we get 12 of these vectors.

3. We ensemble the opinions of the models by taking the average of the 12 vectors per label.

4. As we now have the averaged values for each of the 20 labels we must decide which labels to assign. Therefore we use a threshold. For the values in the vector that are above the threshold, the corresponding label is assigned.

\subsection{Data Preprocessing}
Transformer-based models are trained on large corpora of natural text. Hence, it seems reasonable to transform the input in a format that is most similar to a human-like formulation. Therefore we concatenate the premise, stance and conclusion into a single text string. This transforms the above example (Table \ref{example-argument}) into \textbf{whaling is part of a great number of cultures against We should ban whaling}. 

\subsection{The Models}\label{sec:the models}
The final system is an ensemble of 12 individual models that are based on the transformer architecture. Each model has the same structure: It consists of a  transformer-based language model with one additional fully connected linear layer on top. We use the CLS token as input for the additional layer as described by \citet{Devlin:2018}. The sigmoid function maps the values for each label to the (0,1) interval.

We use two deviations of BERT \cite{Devlin:2018}, namely "microsoft/deberta-large" ("Deberta") and a pretrained roberta-large model ("IBM-Roberta") as base models. 
They are then optimized for loss minimization or f1-score maximization leading to four different configurations as depicted in blue in Figure \ref{fig:architecture}. Each of the four configurations is trained on 3 folds, leading to the 12 models in the final system (Table \ref{model-mapping} in the Appendix). In the following sections, we describe the different configurations in more detail. 
\input{tables/pretraining_before_data_update}

\subsubsection{Pretraining}\label{sec:pretraing}

According to \citet{mirzakhmedova:2023} 83\% of the arguments are retrieved from the IBM-ArgQ-Rank-30kArgs dataset \cite{IBM30k}. We pretrain our models on the IBM-dataset using masked language modeling \cite{Devlin:2018} to shift the language understanding capabilities of our model towards the specific language in the arguments.
Table \ref{pretraining_before_only} shows that pretraining only improves the roberta model but not the deberta model. Subsequently, we proceeded with the pre-trained roberta model (IBM-Roberta) and used the deberta model in its base version.

\subsubsection{Optimizing for Loss and f1-score}

A traditional training procedure requires a train-validation split. To prevent models from over-fitting, the training is stopped as soon as the validation loss reaches its minimum. Within the task, the models are evaluated against the f1-score. So instead of training until the minimum loss is reached we train the models until they reach the maximum f1-score.  

\begin{equation}
f1 = \frac{2*recall_{macro}*precision_{macro}}{{recall_{macro}+precision_{macro}}} 
\label{equation:f1Score}
\end{equation}

The used f1-score (\ref{equation:f1Score}) differs slightly from the f1-scores as defined in most software packages (e.g macro average f1-score in sklearn\footnote{\url{https://scikit-learn.org/stable/modules/generated/sklearn.metrics.f1_score.html}}). Instead of calculating the f1-score for each label and then taking the average, the "macro average recall"\footnote{\url{https://scikit-learn.org/stable/modules/generated/sklearn.metrics.recall_score.html}} and "macro average precision"\footnote{\url{https://scikit-learn.org/stable/modules/generated/sklearn.metrics.precision_score.html}} are calculated first and are then used in the f1-score (formula \ref{equation:f1Score}).

\input{algorithms/threshold_selection.tex}
Now the question arises of how to optimize for the f1-score. During one validation step, we get the predictions for the validation set. Instead of calculating the loss, we could binarize the predictions according to a threshold and use them to calculate the f1-score. We could then train the model until no further improvement in the f1-score is observed. This procedure would require that we select the threshold at each validation step that maximizes the f1-score. 

Consequently, we define Algorithm \ref{alg:optT} to determine the threshold that maximizes the f1-score. As input, it takes a set of true labels and a set of predictions with values between 0 and 1. We iterate over all possible thresholds in small steps (0.01). In each iteration, we binarize the predictions according to the threshold and calculate the corresponding f1-score. As output, we return the threshold that maximizes the f1-score. 

\input{figures/f1-val-figure.tex}

 In Figure \ref{fig:F1ValFigure} we can clearly see, how the model improves its performance during training. It further shows that the training step that minimizes the average validation loss and the training step maximizing the f1-score can differ. The average validation loss (blue) has its minimum at step 3, whereas the f1-score (green) has its maximum at training step 4. Subsequently, we decide to optimize each model with respect to the f1-score and with respect to the loss. This leads to the two blue rows in Figure \ref{fig:architecture}. The first blue row in Figure \ref{fig:architecture} indicates the training with the goal of f1-score maximization, whereas the second row indicates the validation loss minimization.

\subsubsection{Cross-Validation}
To make use of as much training data as possible we apply a variant of 3-fold-cross-validation. Each of the four model configurations is trained three times, with a validation set of 500 samples. Each validation set is created by taking a different random split from the training dataset. Figure \ref{fig:architecture} reflects these 3 versions of each model with the blue boxes in the background.  

\subsection{Ensembling}\label{ensembling}
During the training process, various models (section \ref{sec:the models}) are developed. Hence, the question arises of how to ensemble them to create the final labels. Our final submissions contain two different approaches.

The best-performing system uses Algorithm \ref{alg:optT} to select an optimal threshold that can be used for the test dataset. In order to do so, the models are not trained on the whole dataset but instead, we split a "Leave-Out-Dataset" of 300 samples apart. These 300 samples are not seen by any of the models before and are used to determine the optimal threshold.

\textbf{Recipe I:} 1. Get the predictions on the "Leave-Out-Dataset" for all single models. 2. Average the individual predictions. 3. Select the optimal threshold for the "Leave-Out-Dataset" that maximizes the f1-score with Algorithm \ref{alg:optT} 4. Repeat steps 1 and 2 for the test dataset and use the optimal threshold for the final prediction. 

We also submitted the best-performing system on the "Nahj al-Balagha"-dataset. This system has the same architecture but uses stacking \cite{stacking} as an ensemble method. Instead of defining one "global" threshold for all labels, we train logistic regressions for each label to decide whether a label should be 0 or 1. The models are trained on the entire dataset.

\textbf{Recipe II:} 1. Get the predictions for 3000 samples of the training dataset for all single models. 2. Train multiple logistic regressions\footnote{\url{https://scikit-learn.org/stable/modules/generated/sklearn.multioutput.MultiOutputClassifier.html\#sklearn.multioutput.MultiOutputClassifier}} (input: predictions, output: true labels) to predict the labels based on the predictions. 3. Get the predictions for the test dataset and use the trained logistic regressions to predict the final labels.

\section{Experimental Setup}

The final system was trained on the data provided by the task organizers (training + validation set) except for a "Leave-Out-Dataset" of 300 samples. During the different cross-validation runs a validation set of 500 samples is taken from the training data. We use a linear learning rate schedule and early stopping. We stop the training process if the validation loss or f1-score does not improve in 3 consecutive evaluation steps. Especially the parameter for the learning rate schedule (total\_training\_steps) was manually optimized because it defines the speed at which the learning rate decays and is therefore crucial for the learning process. The hyperparameters for pretraining and finetuning, used model versions as well as links to the code and a docker container reproducing the results can be found in Appendix \ref{app:experiments}.

During the competition, we used an internal test dataset of 500 samples to create an internal "leaderboard" and choose the final systems to submit in the competition.  

The models are implemented with PyTorch-Lightning and are trained on an NVIDIA Tesla T4 GPU. 

\section{Results}

Each team was allowed to make up to four submissions (Table \ref{table-results-combined}).  

\textbf{EN-Thres-LoD}: The system described in this paper and the best-performing system in the competition. The ensemble threshold is calculated on a "Leave-out-Dataset" ("EN-Thres-LoD"), and the model achieves an f1-score of 0.56 (Table \ref{table-results-combined}). 

\textbf{EN-Log-Reg}: A system that uses stacking with logistic regressions as an ensemble method (Section \ref{ensembling}). It has an f1-score of 0.54 on the "Main" dataset and was the best-performing system with an f1-score of 0.40 on the "Nahj al-Balagha" dataset. 

\textbf{EN-Thres-Train}: The same system architecture as EN-Thres-LoD. Instead of calculating the threshold on the "Leave-Out-Dataset", the entire dataset was used for training and the optimal threshold was calculated on the training data. 

\textbf{EN-Silver-Labels}: The system uses a self-training approach (Appendix \ref{app:silver-labels}). A first version of a system creates additional training data ("silver labels") for arguments from the IBM-30k dataset. The model architecture is the same as "EN-Thres-LoD". The size of the training data is then increased to 140\% by adding the "silver-label" data. We followed a similar approach as in \citet{self-training}. The performance during training seems to slightly improve (Figure \ref{app:silver-labels}), but when ensembled the system was outperformed by the other architectures.

All of our four submitted systems rank high in the competition on the "Main" dataset and the "Nahj al-Balagha". On these datasets, our systems outperform the BERT baseline by a large margin. However, there is a different picture for the "New York Times" dataset. With a maximum score of 0.27, the systems are only slightly better than the BERT baseline and are outperformed by other systems by a large margin (0.34 as "best approach" in Table \ref{table-results-combined}).

The scores reported in Table \ref{table-results-combined} refer to the overall scores across all labels. Clearly, the model delivers better results for some labels than for others. A table with the scores for each label can be found in Table \ref{app:results} in the appendix. The system delivers the best performances for the labels "Universalism: nature" and "Security: personal" with f1-scores of 0.82 and 0.76 respectively. The weakest performances are seen for the labels "Hedonism" (0.25) and "Stimulation" (0.32). The dataset is imbalanced and there seems to be a correlation between the frequencies of labels in the training data and the performance of the system (Figure \ref{app:fig:f1_vs_number_samples}).

\input{tables/table_results_combined.tex}

\subsection{Ablation study}
The above-presented approach turned out to be sub-optimal. Based on our own "internal leaderboard" it seemed like the suggested ensemble of all 12 models has slightly superior performance (F1 inter. in Table \ref{ablation-ensemble-comparison} in Appendix \ref{app:internal-leaderboard}). After submission, we evaluated the individual components of the final ensemble on the official test dataset. Table \ref{table-results-combined} shows that ensembling the 3 deberta models, which are optimized for the f1-score (EN-Deberta-F1) leads to slightly better performance (0.57) on the "Main" dataset and significantly better performance on the "New York Times" dataset (0.34). A single Deberta model, optimized for f1-score (Single-Deberta-F1) leads to slightly worse performance on the test dataset (0.55) but significantly better performance on the "New York Times" dataset (0.37). The model selection for these ensembles can be found in Table \ref{model-mapping} in the Appendix.

\subsection{Other approaches}
Besides the different approaches submitted for the competition, we have experimented with the generative T5 transformer model. The T5 model is trained on multiple downstream tasks including natural language inference. We assumed that this multi-tasking ability might have a beneficial influence on our task. Therefore we fine-tuned a T5-large model to predict the values. With an f1-score of 0.493 on the validation set it beats the baseline BERT model but is far from the performance of the other approaches. We further weighted the loss function according to the class distribution in order to account for the class imbalance problem. This was ruled out in the early stages of the system development because it showed slightly lower performance. Comparing several approaches can be complicated due to the randomness in training procedures and the need for different hyperparameter optimizations. These modeling decisions are based on the performances captured in Table \ref{app:table:other_approaches_results} in the appendix.

\section{Conclusion}
We have presented the best-performing system for the task of automatically detecting human values in arguments. Our system ensembles 12 models that have been either optimized for loss minimization or f1-score maximization. As an ensemble method, we choose one global decision threshold for all labels. The threshold maximizes the f1-score for a "Leave-out-Dataset". This approach outperforms stacking as an ensemble method, where for each label a logistic regression is trained. Even though our systems show the best performances on the "Main" and "Nahj al-Balagha" datasets, they are outperformed by other approaches on the "New York Times" dataset. In the ablation study, we show that such a large ensemble is not necessary. In fact, an ensemble of only 3 models shows better performance and robustness while decreasing the system's memory requirements significantly.

For future work, an analysis of this phenomenon could be considered. Does the reduction of ensemble size lead to a more robust system and what are the counter-productive elements in the larger ensemble? Furthermore, we have only manually combined a few different models into an ensemble. Hence, it would be interesting to see whether a systematic selection of different approaches within an ensemble could further boost performance.

\section{Limitations}
The proposed system is trained on a very specific argument structure taken from the IBM-30k dataset. The system's performance noticeably declines when tested on additional datasets, which raises questions about their ability to handle new, unprepared datasets and textual arguments with robustness. Furthermore, the best performing system consists of 12 individual models leading to a high resource requirements. 

\section{Acknowledgments}
We would like to thank Nicolas Handke and Johannes Kiesel for creating the docker container and the web application and thereby making the system easier accessible for future research.

\bibliography{custom}
\bibliographystyle{acl_natbib}

\appendix
\section{Experimental Setup}\label{app:experiments}
This section includes the appendices for the experimental setup such as links to relevant resources and hyperparameters.
\subsection{Code, Docker Container and Models}
The Code\footnote{\url{https://github.com/danielschroter/human_value_detector}} and Docker Container\footnote{\url{https://github.com/touche-webis-de/team-adam-smith23}} for the final system are available online.
We pushed the Single Deberta Model\footnote{\url{https://huggingface.co/tum-nlp/Deberta_Human_Value_Detector}} from the ablation studies to Huggingface for simple usage. 
Furthermore, all models are publicly available.\footnote{\url{https://zenodo.org/record/7656534\#.Y_yKdyaZP30}} 

\subsection{Language Models}
We used the microsoft/deberta-large\footnote{\url{https://huggingface.co/microsoft/deberta-large}} model from huggingface and the pretrained roberta-large.\footnote{\url{https://huggingface.co/danschr/roberta-large-BS_16-EPOCHS_8-LR_5e-05-ACC_GRAD_2-MAX_LENGTH_165/tree/main?doi=true}}

\subsection{Hyperparameters}

\input{tables/Hyperparams}

\input{tables/hp-pretraining}

\section{Results}\label{app:results}

\subsection{Self-training and Silver Labels} \label{app:silver-labels}
We further applied a self-training procedure: 1)~First we trained the IBM-Roberta and Deberta Model for f1-score-maximization. 2)~We then ensembled the models and defined the optimal threshold based on a small Leave\_out\_Dataset. 3)~We used the ensemble to created additional silver labels (additional training data) from the IBM-30k dataset. Thereby we ensured to not include samples, that are in our internal test dataset or the test dataset of the competition. 4)~We retrained the models on an extended dataset including the silver labels in the training data. (During Cross-Validation, it is ensured that silver-labels are not added to the validation set). 
Figure \ref{fig:silverlabels} shows how pretraining might have a positive impact on the performance of the model. The values are the averages from cross-validating with 3 folds. However, the submitted system (ensemble of 12) containing the self-trained models was outperformed by the other models. 

\input{figures/silver_labels}


\subsection{Results per category}
\input{tables/table_results}
Table \ref{table-results} contains the performances of the systems submitted for the competition and developed in the ablation studies. It shows the performance for each label individually.

\begin{figure*}
    \centering
    \includegraphics[width=.7\linewidth]{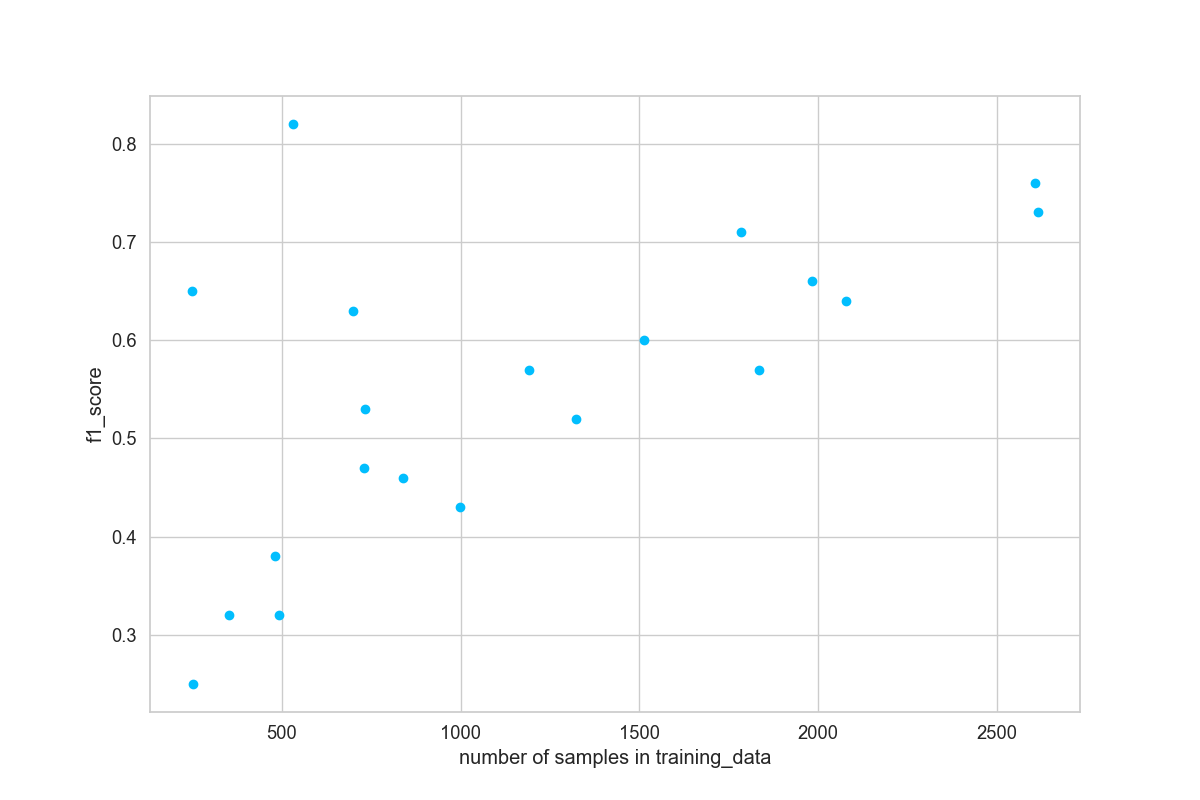}\hfill
    \caption{Error Analysis: Label Frequency in training data vs f1-performance of the system.}
    \label{app:fig:f1_vs_number_samples}
\end{figure*}
\input{tables/other_approaches_results}
\newpage
\subsection{Internal Leaderboard for Ensembling}\label{app:internal-leaderboard}
Table \ref{ablation-ensemble-comparison} compares the performance of the different ensembles. Based on the f1-score performance on an internal test-split we decided to submit the system with 12 models.  
\input{tables/ablation-ensemble-comparison.tex}

\input{tables/Model_Mapping}
\subsection{Error Analysis}
We plotted the frequencies of the labels in the training data against their f1-score performance (Figure \ref{app:fig:f1_vs_number_samples})

\subsection{Other Approaches}
We provide the results of some different approaches. They have been calculated on the same validation set of 500 samples, but we initialized the training with three different random seeds. Table \ref{app:table:other_approaches_results} contains different methodologies.

\subsection{Model Mapping - Ablation Studies}
Table \ref{model-mapping} shows the identifier of the models in the model repository together with optimization goals and random seeds. The Table also shows which model is included in the Ensembles in the Ablation Studies.

\end{document}

%% file: tables/example.tex
\begin{table}[H]
\centering\small%
\setlength{\tabcolsep}{2.5pt}%
\begin{tabular}{@{}ll@{\hspace{10pt}}c@{\hspace{5pt}}cccccccccccccccccccccc@{}}
\midrule
& Premise   & whaling is part of a great number of cultures\\
& Conclusion  & We should ban whaling \\
& Stance   & against \\
& Labels & [’Tradition’, ’Conformity: interpersonal’] \\
\midrule

\end{tabular}
\caption{Example argument about whaling}
\label{example-argument}
\end{table}

%% file: tables/pretraining_before_data_update.tex
\begin{table}[h]
\centering\small%
\setlength{\tabcolsep}{2.5pt}%
\begin{tabular}{@{}ll@{\hspace{10pt}}c@{\hspace{5pt}}cccccccccccccccccccccc@{}}
\toprule
\multicolumn{2}{@{}l}{\bf Pretraining} & \bf  {\bf F1 Validation}  \\
\midrule
& IBM-Deberta-Large   & .516 \\
& Microsoft Deberta Large  & \bf{.523} \\

& IBM-Roberta-large   & \bf{.529} \\
& Roberta-large  & .519 \\

\bottomrule
\end{tabular}
\caption{Pretraining: Values are calculated without tuned hyperparameters and before the training data update by the organizers during the competition.}
\label{pretraining_before_only}
\end{table}

%% file: algorithms/threshold_selection.tex
\begin{algorithm}[t]
\footnotesize
	\caption{Threshold selection with f1-score maximization} 
        \label{alg:optT}
	\begin{algorithmic}[1]
        \State Input: yTrue, yPred
        \State Output: optimal threshold optT
        \State
        \State $threshold \gets 0$
        \State $maxF1 \gets 0$
        \State $optT \gets 0$
            \While{$threshold \leq 1$}
                \State $yPredBin \gets$ binarize yPred with $threshold$
                \State $recall \gets $ recall(yTrue, yPredBin, average="macro")
                \State $precision \gets $ precision(yTrue, yPredBin, \newline \hspace*{11em}average="macro")
                \If {$precision+recall \neq 0$}
                    \State $f1 \gets \displaystyle{\dfrac{2*recall*precision}{(recall+precision)}}$
                    \If{$f1 \geq maxF1$}
                        \State $optT \gets threshold$
                        \State $maxF1 \gets f1$
                    \EndIf
                \EndIf
                \State $threshold \gets threshold + 0.01$ 
            \EndWhile
        \State return $optT$
	\end{algorithmic} 
\end{algorithm}

%% file: figures/f1-val-figure.tex
\begin{figure}[h]
    \centering
    \noindent\resizebox{\linewidth}{!}{%
    \begin{tikzpicture}
    legend columns = 4,
    legend entries {{\tiny Micro F1}{\tiny Macro F1}{\tiny Custom F1} {\tiny Loss}},
    legend style={at=(0.5, -0.1)}, anchor=north,
    legend to name=testLegend
    \begin{axis}[
        xlabel=Training Steps,
        ylabel={F1-Scores},
        xmin=0, xmax=8,
        ymin=0.45, ymax=0.65,
        axis y line*=left,
        xtick={0,1,2,3,4,5,6,7,8},
        ytick={0.36,0.40,0.44,0.48,0.52,0.56,0.60,0.64},
        ymajorgrids=true,
        grid style=dashed,
    ]
    \addplot[
        color=red,
        mark=square]
        coordinates{
        (0,0.4538)(1,0.5821)(2,0.6172)(3,0.6281)(4,0.6365)(5,0.6356)(6,0.6285)(7,0.6285)
        };
        \label{pgfplots:microF1}
    \addplot[
        color=orange,
        mark=square]
        coordinates{
        (0,0.3901)(1,0.4954)(2,0.5477)(3,0.5649)(4,0.5878)(5,0.5828)(6,0.5850)(7,0.5850)
        };
        \label{pgfplots:macroF1}
    \addplot[
        color=green,
        mark=square]
        coordinates{
        (0,0.4481)(1,0.5264)(2,0.5796)(3,0.5746)(4,0.5975)(5,0.5883)(6,0.5889)(7,0.5889)
        };
        \label{pgfplots:customF1}
    \end{axis}
    \begin{axis}[
        legend columns=2,
        legend to name=named,
        xmin=0, xmax=8,
        ymin=0.28, ymax=0.35,
        axis y line*=right,
        xtick={0,1,2,3,4,5,6,7,8},
        ytick={0.28,0.29,0.30,0.31,0.32,0.33,0.34,0.35},
        ylabel=Average Val Loss,
        ylabel near ticks
    ]
    \addlegendimage{/pgfplots/refstyle=pgfplots:microF1}\addlegendentry{\tiny Micro F1}
    \addlegendimage{/pgfplots/refstyle=pgfplots:macroF1}\addlegendentry{\tiny Macro F1}
    \addlegendimage{/pgfplots/refstyle=pgfplots:customF1}\addlegendentry{\tiny F1 (\ref{equation:f1Score})}
    \addplot[
        color=blue,
        mark=diamond]
        coordinates{
        (0,0.3496)(1,0.3120)(2,0.2907)(3,0.2868)(4,0.2884)(5,0.2869)(6,0.2867)(7,0.2867)
        };
        \addlegendentry{\tiny avgValLoss}
    
    \end{axis} 
    \end{tikzpicture}
    }
    \ref{named}
    \caption{F1-scores and average validation loss in each training step in a single run of a Deberta model}
    \label{fig:F1ValFigure}

\end{figure}

%% file: tables/table_results_combined.tex
\begin{table}
\centering\small%
\setlength{\tabcolsep}{2.5pt}%
\begin{tabular}{@{}ll@{\hspace{10pt}}c@{\hspace{5pt}}cccccccccccccccccccccc@{}}
\toprule
\multicolumn{2}{@{}l}{\bf Approach / Test Dataset} & \rotatebox{90}{\bf Main} & \rotatebox{90}{\bf Nahj al-Balagha} & \rotatebox{90}{\bf New York Times} & \rotatebox{90}{Threshold} & \rotatebox{90}{\# models} \\
\midrule
\multicolumn{2}{@{}l}{\emph{Main}} \\
& \textcolor{gray}{Best per category} & \textcolor{gray}{.59} & \textcolor{gray}{.48} & \textcolor{gray}{.47}\\
& \textcolor{gray}{Best approach} & \textcolor{gray}{.56} & \textcolor{gray}{.40}  & \textcolor{gray}{.34} \\
& \textcolor{gray}{BERT} & \textcolor{gray}{.42} & \textcolor{gray}{.28}  & \textcolor{gray}{.24} \\
& \textcolor{gray}{1-Baseline} & \textcolor{gray}{.26} & \textcolor{gray}{.13} & \textcolor{gray}{.15} \\
\multicolumn{2}{@{}l}{\emph{Submitted Models}} \\
& EN-Thres-Train  & .56 & .36 & .26 & 0.32 & 12\\
& EN-Log-Reg  & .54 & \bf{.40} & \bf{.27} & - & 12\\
& EN-Thres-LoD (1st) & \bf{.56} & .34 & .25 & 0.26  & 12 \\
& EN-Silver-Labels & .54 & .34  & .24 & 0.29 & 12\\

\multicolumn{2}{@{}l}{\emph{Ablation Studies}} \\
& EN-Deberta-F1* & .57 & .33 & .34 & 0.27 & 3\\
& Single-Deberta-F1* &.55 & .35 & .37 & 0.25 & 1\\

\bottomrule
\end{tabular}
\caption{Achieved f1-score of team adam-smith per test dataset. Approaches marked with * were not part of the official evaluation. Approaches in gray are shown for comparison: an ensemble using the best participant approach for each individual category; the best participant approach; and the organizer's BERT and 1-Baseline.}
\label{table-results-combined}
\end{table}

%% file: tables/Hyperparams.tex
\begin{table}[H]
\centering\small%
\setlength{\tabcolsep}{2.5pt}%
\begin{tabular}{@{}ll@{\hspace{10pt}}c@{\hspace{5pt}}cccccccccccccccccccccc@{}}
\toprule
\multicolumn{2}{@{}l}{\bf Parameters} & {\bf Value} & \\
\midrule
\multicolumn{2}{@{}l}{\emph{General}} \\
& batch\_size & 8 \\
& epochs* & 3* (see caption) \\

\multicolumn{2}{@{}l}{\emph{Optimizer}} \\
& Optimzier & AdamW\\
& learning\_rate\_schedule & linear \\
& Learning Rate  & 2e-5\\
& total\_training\_steps  & 2502 \\
& n\_warmup\_steps  & 500  \\

\multicolumn{2}{@{}l}{\emph{Early\_Stopping}} \\
& validation\_inverval & 300 \\
& epochs & early\_stopping* \\
& patience & 3\\

\addlinespace

\bottomrule
\end{tabular}
\caption{Hyperparameters: The epochs are not actually the number of trained epochs. Instead they are used to calculate the linear learning rate schedule by calculating the Total\_training\_steps = (len(training\_data)//Batch\_Size)*Epochs. The models are then trained with early stopping }
\label{hyperparameters}
\end{table}

%% file: tables/hp-pretraining.tex
\begin{table}[H]
\centering\small%
\setlength{\tabcolsep}{2.5pt}%
\begin{tabular}{@{}ll@{\hspace{10pt}}c@{\hspace{5pt}}cccccccccccccccccccccc@{}}
\toprule
\multicolumn{2}{@{}l}{\bf Parameters} & {\bf Value} & \\
\midrule
& Batch Size & 16 \\
& accumulated gradients & 2 \\
& epochs & 8 \\
& Learning Rate  & 2e-5\\

\bottomrule
\end{tabular}
\caption{Hyperparameters Pretraining with Masked Language Modelling}
\label{hp_pretraining}
\end{table}

%% file: figures/silver_labels.tex
\begin{figure}[h]
    \centering
    \noindent\resizebox{\linewidth}{!}{%
    \begin{tikzpicture}
    \begin{axis}[
        xlabel=Fraction of training data.,
        ylabel={F1-Scores},
        xmin=0.8, xmax=1.8,
        ymin=0.52, ymax=0.61,
        xtick={0.8,1,1.2,1.4,1.6},
        ytick={0.51,0.52,0.52,0.53,0.54,0.55,0.56,0.57, 0.58, 0.59,0.60,0.61},
        ymajorgrids=true,
        grid style=dashed,
        legend pos= south east,
    ]
    \addplot[
        color=red,
        mark=square]
        coordinates{
        (0.8,0.529)(1,0.553)(1.2,0.572)(1.4,0.558)(1.6,0.561)
        };
        \label{pgfplots:F1Test}
    \addplot[
        color=orange,
        mark=square]
        coordinates{
        (0.8,0.593)(1,0.581)(1.2,0.590)(1.4,0.595)(1.6,0.604) 
        };
        \label{pgfplots:F1Val}

    \addlegendimage{/pgfplots/refstyle=pgfplots:F1Test}\addlegendentry{\tiny F1Test}
    \addlegendimage{/pgfplots/refstyle=pgfplots:F1Val}\addlegendentry{\tiny F1Val}

    \end{axis}
    \end{tikzpicture}
    }
    \caption{F1 Scores on average Validation loss during Cross-Validation (F1 Val) and the internal Test-Dataset (F1 Test).
    The X Axis represents the amount of training data used. Values above 1 indicate that additional silver-labels are included in the training.}
    \label{fig:silverlabels}

\end{figure}

%% file: tables/table_results.tex
\begin{table*}
\centering\small%
\setlength{\tabcolsep}{2.5pt}%
\begin{tabular}{@{}ll@{\hspace{10pt}}c@{\hspace{5pt}}cccccccccccccccccccccc@{}}
\toprule
\multicolumn{2}{@{}l}{\bf Test set / Approach} & \bf All & \rotatebox{90}{\bf Self-direction: thought} & \rotatebox{90}{\bf Self-direction: action} & \rotatebox{90}{\bf Stimulation} & \rotatebox{90}{\bf Hedonism} & \rotatebox{90}{\bf Achievement} & \rotatebox{90}{\bf Power: dominance} & \rotatebox{90}{\bf Power: resources} & \rotatebox{90}{\bf Face} & \rotatebox{90}{\bf Security: personal} & \rotatebox{90}{\bf Security: societal} & \rotatebox{90}{\bf Tradition} & \rotatebox{90}{\bf Conformity: rules} & \rotatebox{90}{\bf Conformity: interpersonal} & \rotatebox{90}{\bf Humility} & \rotatebox{90}{\bf Benevolence: caring} & \rotatebox{90}{\bf Benevolence: dependability} & \rotatebox{90}{\bf Universalism: concern} & \rotatebox{90}{\bf Universalism: nature} & \rotatebox{90}{\bf Universalism: tolerance} & \rotatebox{90}{\bf Universalism: objectivity} \\
\midrule
\multicolumn{2}{@{}l}{\emph{Main}} \\
& \textcolor{gray}{Best per category} & \textcolor{gray}{.59} & \textcolor{gray}{.61} & \textcolor{gray}{.71} & \textcolor{gray}{.39} & \textcolor{gray}{.39} & \textcolor{gray}{.66} & \textcolor{gray}{.50} & \textcolor{gray}{.57} & \textcolor{gray}{.39} & \textcolor{gray}{.80} & \textcolor{gray}{.68} & \textcolor{gray}{.65} & \textcolor{gray}{.61} & \textcolor{gray}{.69} & \textcolor{gray}{.39} & \textcolor{gray}{.60} & \textcolor{gray}{.43} & \textcolor{gray}{.78} & \textcolor{gray}{.87} & \textcolor{gray}{.46} & \textcolor{gray}{.58} \\
& \textcolor{gray}{Best approach} & \textcolor{gray}{.56} & \textcolor{gray}{.57} & \textcolor{gray}{.71} & \textcolor{gray}{.32} & \textcolor{gray}{.25} & \textcolor{gray}{.66} & \textcolor{gray}{.47} & \textcolor{gray}{.53} & \textcolor{gray}{.38} & \textcolor{gray}{.76} & \textcolor{gray}{.64} & \textcolor{gray}{.63} & \textcolor{gray}{.60} & \textcolor{gray}{.65} & \textcolor{gray}{.32} & \textcolor{gray}{.57} & \textcolor{gray}{.43} & \textcolor{gray}{.73} & \textcolor{gray}{.82} & \textcolor{gray}{.46} & \textcolor{gray}{.52} \\
& \textcolor{gray}{BERT} & \textcolor{gray}{.42} & \textcolor{gray}{.44} & \textcolor{gray}{.55} & \textcolor{gray}{.05} & \textcolor{gray}{.20} & \textcolor{gray}{.56} & \textcolor{gray}{.29} & \textcolor{gray}{.44} & \textcolor{gray}{.13} & \textcolor{gray}{.74} & \textcolor{gray}{.59} & \textcolor{gray}{.43} & \textcolor{gray}{.47} & \textcolor{gray}{.23} & \textcolor{gray}{.07} & \textcolor{gray}{.46} & \textcolor{gray}{.14} & \textcolor{gray}{.67} & \textcolor{gray}{.71} & \textcolor{gray}{.32} & \textcolor{gray}{.33} \\
& \textcolor{gray}{1-Baseline} & \textcolor{gray}{.26} & \textcolor{gray}{.17} & \textcolor{gray}{.40} & \textcolor{gray}{.09} & \textcolor{gray}{.03} & \textcolor{gray}{.41} & \textcolor{gray}{.13} & \textcolor{gray}{.12} & \textcolor{gray}{.12} & \textcolor{gray}{.51} & \textcolor{gray}{.40} & \textcolor{gray}{.19} & \textcolor{gray}{.31} & \textcolor{gray}{.07} & \textcolor{gray}{.09} & \textcolor{gray}{.35} & \textcolor{gray}{.19} & \textcolor{gray}{.54} & \textcolor{gray}{.17} & \textcolor{gray}{.22} & \textcolor{gray}{.46} \\
& EN-Thres-Train  & .56 & .59 & \bf{.71} & .22 & .29 & \bf{.66} & .48 & .52 & .30 & \bf{.79} & .67 & \bf{.65} & .61 & .61 & .19 & \bf{.60} & .36 & .74 & .84 & .41 & \bf{.53} \\
& EN-Log-Reg  & .54 & \bf{.61} & \bf{.71} & .20 & .29 & .62 & .46 & .44 & .30 & .78 & \bf{.68} & .64 & .59 & .61 & .20 & .59 & .36 & \bf{.76} & \bf{.85} & .38 & .49 \\
& EN-Thres-LoD (1st) & .56 & .57 & \bf{.71} & \bf{.32} & .25 & \bf{.66} & .47 & .53 & \bf{.38} & .76 & .64 & .63 & .60 & \bf{.65} & \bf{.32} & .57 & \bf{.43} & .73 & .82 & .46 & .52 \\
& EN-Silver-Labels & .54 & .58 & .70 & .13 & .29 & .65 & .45 & .53 & .19 & .73 & .59 & .64 & .55 & .60 & .16 & .57 & .38 & .71 & .84 & .46 & .50 \\
& EN-Deberta-F1* & \bf{.57} & .57 & \bf{.71} & .30 & \bf{.34} & .65 & \bf{.50} & \bf{.55} & \bf{.38} & .78 & .64 & .64 & .60 & .60 & \bf{.32} & .57 & \bf{.43} & .75 & .83 & \bf{.47} & \bf{.53}\\
& Single-Deberta-F1* &.55 & .54 & .70 & .29 & .32 & .65 & .44 & .55 & .37 & .77 & .63 & .62 & \bf{.62} & \bf{.65} & .29 & .55 & .42 & .74 & .81 & .46 &.52\\
\addlinespace
\multicolumn{2}{@{}l}{\emph{Nahj al-Balagha}} \\
& \textcolor{gray}{Best per category} & \textcolor{gray}{.48} & \textcolor{gray}{.18} & \textcolor{gray}{.49} & \textcolor{gray}{.50} & \textcolor{gray}{.67} & \textcolor{gray}{.66} & \textcolor{gray}{.29} & \textcolor{gray}{.33} & \textcolor{gray}{.62} & \textcolor{gray}{.51} & \textcolor{gray}{.37} & \textcolor{gray}{.55} & \textcolor{gray}{.36} & \textcolor{gray}{.27} & \textcolor{gray}{.33} & \textcolor{gray}{.41} & \textcolor{gray}{.38} & \textcolor{gray}{.33} & \textcolor{gray}{.67} & \textcolor{gray}{.20} & \textcolor{gray}{.44} \\
& \textcolor{gray}{Best approach} & \textcolor{gray}{.40} & \textcolor{gray}{.13} & \textcolor{gray}{.49} & \textcolor{gray}{.40} & \textcolor{gray}{.50} & \textcolor{gray}{.65} & \textcolor{gray}{.25} & \textcolor{gray}{.00} & \textcolor{gray}{.58} & \textcolor{gray}{.50} & \textcolor{gray}{.30} & \textcolor{gray}{.51} & \textcolor{gray}{.28} & \textcolor{gray}{.24} & \textcolor{gray}{.29} & \textcolor{gray}{.33} & \textcolor{gray}{.38} & \textcolor{gray}{.26} & \textcolor{gray}{.67} & \textcolor{gray}{.00} & \textcolor{gray}{.36} \\
& \textcolor{gray}{BERT} & \textcolor{gray}{.28} & \textcolor{gray}{.14} & \textcolor{gray}{.09} & \textcolor{gray}{.00} & \textcolor{gray}{.67} & \textcolor{gray}{.41} & \textcolor{gray}{.00} & \textcolor{gray}{.00} & \textcolor{gray}{.28} & \textcolor{gray}{.28} & \textcolor{gray}{.23} & \textcolor{gray}{.38} & \textcolor{gray}{.18} & \textcolor{gray}{.15} & \textcolor{gray}{.17} & \textcolor{gray}{.35} & \textcolor{gray}{.22} & \textcolor{gray}{.21} & \textcolor{gray}{.00} & \textcolor{gray}{.20} & \textcolor{gray}{.35} \\
& \textcolor{gray}{1-Baseline} & \textcolor{gray}{.13} & \textcolor{gray}{.04} & \textcolor{gray}{.09} & \textcolor{gray}{.01} & \textcolor{gray}{.03} & \textcolor{gray}{.41} & \textcolor{gray}{.04} & \textcolor{gray}{.03} & \textcolor{gray}{.23} & \textcolor{gray}{.38} & \textcolor{gray}{.06} & \textcolor{gray}{.18} & \textcolor{gray}{.13} & \textcolor{gray}{.06} & \textcolor{gray}{.13} & \textcolor{gray}{.17} & \textcolor{gray}{.12} & \textcolor{gray}{.12} & \textcolor{gray}{.01} & \textcolor{gray}{.04} & \textcolor{gray}{.14} \\
& EN-Thres-Train & .36 & .12 & .43 & \bf{.50} & \bf{.50} & \bf{.66} & .22 & .00 & .56 & .50 & .23 & .55 & .23 & .15 & \bf{.31} & .30 & .27 & .26 & .40 & .00 & .35 \\
& EN-Log-Reg (1st) & \bf{.40} & \bf{.13} & \bf{.49} & .40 & \bf{.50} & .65 & {.25} & .00 & .58 & .50 & \bf{.30} & .51 & \bf{.28} & \bf{.24} & .29 & \bf{.33} & \bf{.38} & .26 & \bf{.67} & .00 & \bf{.36} \\
& EN-Thres-LoD & .34 & .09 & .33 & .33 & .44 & .59 & .22 & .20 & \bf{.62} & .51 & .20 & \bf{.55} & .23 & .12 & .24 & .26 & .24 & \bf{.29} & .40 & .05 & .30 \\
& EN-Silver-Labels & .34 & .06 & .37 & .33 & .40 & .62 & .22 & .18 & .51 & .49 & .23 & .51 & .21 & .23 & .20 & .24 & .24 & .24 & .50 & .00 & .32 \\
& EN-Deberta-F1* & .33 & \bf{.13} & .34 & .25 & .31 & .64 & .21 & \bf{.22} & .57 & .53 & .21 & \bf{.55} & .23 &.15 & .27 & .27 & .21 & .24 & .40 & \bf{.11} & .30\\
& Single-Deberta-F1* & .35 & .10 & .35 & .15 & .29 & .65 & .22 & .15 & .55 & \bf{.54} & .25 & .46 & .24 & .17 & .20 & .25 & .23 & .25 & \bf{.67} & .10 & .34\\
\addlinespace
\multicolumn{2}{@{}l}{\emph{New York Times}} \\
& \textcolor{gray}{Best per category} & \textcolor{gray}{.47} & \textcolor{gray}{.50} & \textcolor{gray}{.22} & \textcolor{gray}{-} & \textcolor{gray}{.03} & \textcolor{gray}{.54} & \textcolor{gray}{.40} & \textcolor{gray}{-} & \textcolor{gray}{.50} & \textcolor{gray}{.59} & \textcolor{gray}{.52} & \textcolor{gray}{-} & \textcolor{gray}{.33} & \textcolor{gray}{1.0} & \textcolor{gray}{.57} & \textcolor{gray}{.33} & \textcolor{gray}{.40} & \textcolor{gray}{.62} & \textcolor{gray}{1.0} & \textcolor{gray}{.03} & \textcolor{gray}{.46} \\
& \textcolor{gray}{Best approach} & \textcolor{gray}{.34} & \textcolor{gray}{.22} & \textcolor{gray}{.22} & \textcolor{gray}{-} & \textcolor{gray}{.00} & \textcolor{gray}{.48} & \textcolor{gray}{.40} & \textcolor{gray}{-} & \textcolor{gray}{.00} & \textcolor{gray}{.53} & \textcolor{gray}{.44} & \textcolor{gray}{-} & \textcolor{gray}{.18} & \textcolor{gray}{1.0} & \textcolor{gray}{.20} & \textcolor{gray}{.12} & \textcolor{gray}{.29} & \textcolor{gray}{.55} & \textcolor{gray}{.33} & \textcolor{gray}{.00} & \textcolor{gray}{.36} \\
& \textcolor{gray}{BERT} & \textcolor{gray}{.24} & \textcolor{gray}{.00} & \textcolor{gray}{.00} & \textcolor{gray}{-} & \textcolor{gray}{.00} & \textcolor{gray}{.29} & \textcolor{gray}{.00} & \textcolor{gray}{-} & \textcolor{gray}{.00} & \textcolor{gray}{.53} & \textcolor{gray}{.43} & \textcolor{gray}{-} & \textcolor{gray}{.00} & \textcolor{gray}{.00} & \textcolor{gray}{.57} & \textcolor{gray}{.26} & \textcolor{gray}{.27} & \textcolor{gray}{.36} & \textcolor{gray}{.50} & \textcolor{gray}{.00} & \textcolor{gray}{.32} \\
& \textcolor{gray}{1-Baseline} & \textcolor{gray}{.15} & \textcolor{gray}{.05} & \textcolor{gray}{.03} & \textcolor{gray}{-} & \textcolor{gray}{.03} & \textcolor{gray}{.28} & \textcolor{gray}{.03} & \textcolor{gray}{-} & \textcolor{gray}{.05} & \textcolor{gray}{.51} & \textcolor{gray}{.20} & \textcolor{gray}{-} & \textcolor{gray}{.07} & \textcolor{gray}{.03} & \textcolor{gray}{.12} & \textcolor{gray}{.12} & \textcolor{gray}{.26} & \textcolor{gray}{.24} & \textcolor{gray}{.03} & \textcolor{gray}{.03} & \textcolor{gray}{.33} \\
&  EN-Thres-Train & .26 & .29 & .14 & - & .00 & \bf{.54} & .00 & - & .00 & .56 & .42 & - & .23 & .00 & .00 & \bf{.33} & \bf{.40} & .58 & .33 & .00 & .40 \\
& EN-Log-Reg & .27 & \bf{.33} & \bf{.18} & - & .00 & .42 & .00 & - & .00 & \bf{.58} & \bf{.52} & - & .18 & .00 & .00 & .21 & .31 & \bf{.62} & \bf{.50} & .00 & \bf{.46} \\
& EN-Thres-LoD & .25 & .18 & .17 & - & .00 & .42 & .00 & - & .00 & .57 & .38 & - & .27 & .00 & .20 & .26 & .37 & .50 & .33 & .00 & .42 \\
& EN-Silver-Labels & .24 & .22 & .15 & - & .00 & .42 & .00 & - & .00 & .56 & .36 & - & \bf{.29} & .00 & .00 & .26 & .34 & .50 & .40 & .00 & .44 \\
& EN-Deberta-F1* & .34 & .22 & .15 & - & \bf{1.0} & .44 & .00 & - & .33 & .56 & .37 & - & \bf{.29} & .00 & .22 & .28 & .24& .44 & .33 & .00 & .42 \\
& Single-Deberta-F1* & \bf{.37} & .22 & .00 & - & \bf{1.0} & .42 & .00 & - & \bf{.40} & .56 & .36 & - & .22 & \bf{.67} & .22 & .21 & .29 & .48 & .33 & .00 & .45 \\
\bottomrule
\end{tabular}
\caption{Achieved f1-score of team adam-smith per test dataset, from macro-precision and macro-recall (All) and for each of the 20~value categories. Approaches marked with * were not part of the official evaluation. Approaches in gray are shown for comparison: an ensemble using the best participant approach for each individual category; the best participant approach; and the organizer's BERT and 1-Baseline. The bold values highlight the best per category approach.}
\label{table-results}
\end{table*}

%% file: tables/other_approaches_results.tex
\begin{table*}[h!]
\centering\small%
\setlength{\tabcolsep}{2.5pt}%
\begin{tabular}{@{}ll@{\hspace{10pt}}c@{\hspace{5pt}}cccccccccccccccccccccc@{}}
\toprule
\multicolumn{2}{@{}l}{\bf Pretraining} & \bf  {\bf F1 Validation}  & \bf{pre-training} & \bf{weighted-loss} &\bf{epochs*} & \bf{optimized} & \bf{LR} & \bf{batch}\\
\midrule
& IBM-Roberta-large   & \bf{.529} & (+) &  & 20 & F1 & 2e-5 & 8 \\
& IBM-Roberta-large & .526 & (+) & (+) & 20 & F1 & 2e-5 & 8 \\
& Roberta-large  & .519 & &  & 20 & F1 & 2e-5 & 8 \\
& T5-large & .493 & & & 35 & F1 & .001 & 16 \\

\bottomrule
\end{tabular}
\caption{Results of other approaches: Values are calculated without tuned hyperparameters.  This training was at the beginning of the competition where instead of applying 3-fold-cross validation we took a validation sample of 500 and trained the model with 3 different random seed initialization. For weighting the loss function we used "Inverse Number of Samples"\footnote{\url{https://medium.com/gumgum-tech/handling-class-imbalance-by-introducing-sample-weighting-in-the-loss-function-3bdebd8203b4}} as weights. The epochs are not actually the number of trained epochs. Instead they are used to calculate the linear learning rate schedule by calculating the Total\_training\_steps = (len(training\_data)//Batch\_Size)*Epochs. The models are then trained with early stopping.}
\label{app:table:other_approaches_results}
\end{table*}

%% file: tables/ablation-ensemble-comparison.tex
\begin{table}[H]
\centering\small%
\setlength{\tabcolsep}{2.5pt}%
\begin{tabular}{@{}ll@{\hspace{10pt}}c@{\hspace{5pt}}cccccccccccccccccccccc@{}}
\toprule
\multicolumn{2}{@{}l}{\bf Model Selection} & {\bf \#} & {\bf Thres.} & {\bf F1 Test} & {\bf F1 inter.}\\
\midrule

& EN-Max-F1*  & 6 & .26 & .555 & .596 \\
& EN-Thres-LoD (1st)  & 12 & .26 & .561 & \bf{.599} \\
& EN-Deberta-F1*  & 3 & .27 & \bf{.566} & .589 \\
& Single-Deberta-F1*  & 1 & .25 &.554 & .565 \\

\addlinespace

\bottomrule
\end{tabular}
\caption{Ablation Studies: Scores on official test set (F1-Test) and scores for internal test split. For Single-Deberta-F1 the Model with Random Seed $= 123$ was selected. \# represents the number of models in ensemble. The models with their identifier are listed in Table \ref{model-mapping} in the Appendix}
\label{ablation-ensemble-comparison}
\end{table}

%% file: tables/Model_Mapping.tex
\begin{table*}[t!]
\centering\small%
\setlength{\tabcolsep}{2.5pt}%
\begin{tabular}{@{}ll@{\hspace{10pt}}c@{\hspace{5pt}}cccccccccccccccccccccc@{}}
\toprule
\multicolumn{2}{@{}l}{\bf Model Mapping} & {\bf{model}} & {\bf Fold-Seed} & {\bf Optimized} & {\bf EN-Max-F1*} & {\bf EN-Deberta-F1*}  & {\bf Single-F1*}\\
\midrule

& HCV-406 & deberta & 42 & F1 & (+) & (+)& &\\
& HCV-408 & deberta &96 & F1 &(+) & (+)& &\\
& HCV-409 & deberta &123 & F1 &(+)& (+)& (+)&\\
& HCV-402 & danschr-roberta &42 & F1 &(+)& &\\
& HCV-403 & danschr-roberta &96 & F1 &(+)& &\\
& HCV-405 & danschr-roberta &123 & F1 &(+)& &\\
& HCV-364 & deberta & 42 & Loss & & & &\\
& HCV-366 & deberta &96 & Loss & & & &\\
& HCV-368 & deberta &123 & Loss & & & &\\
& HCV-371 & danschr-roberta &42 & Loss & & & &\\
& HCV-372 & danschr-roberta &96 & Loss  & & & &\\
& HCV-375 & danschr-roberta &123 & Loss & & & &\\

\bottomrule
\end{tabular}
\caption{The models are publicly available (Appendix A). The danschr-roberta represents the "IBM-Roberta" in the paper. The first column identifies the model in the model repository.}
\label{model-mapping}
\end{table*}